\documentclass{IEEEtran}

\usepackage{subfigure}
\usepackage{cite}
\usepackage{amsmath,amssymb,amsfonts}
\usepackage[ruled,linesnumbered]{algorithm2e}

\usepackage{color}
\usepackage{graphicx}
\usepackage{textcomp}
\usepackage{booktabs}
\usepackage{multirow}
\usepackage{bbding}
\usepackage{bm}
\usepackage{orcidlink}
\def\BibTeX{{\rm B\kern-.05em{\sc i\kern-.025em b}\kern-.08em
	T\kern-.1667em\lower.7ex\hbox{E}\kern-.125emX}}

\begin{document}

\title{
    \vspace{-1.5cm}
    \textnormal{
        \begin{minipage}{\textwidth}
            \centering
            \small
            This work has been submitted to the IEEE for possible publication. Copyright may be transferred without notice, after which this version may no longer be accessible.
        \end{minipage}
    } \\
    \vspace{0.8cm} 
    TS-MLLM: A Multi-Modal Large Language Model-based Framework for Industrial Time-Series Big Data Analysis
}

\author{
{
Haiteng Wang~\orcidlink{0000-0002-7316-3607},~\IEEEmembership{Graduate Student Member,~IEEE,}
Yikang Li~\orcidlink{0009-0005-7459-8660},~\IEEEmembership{Graduate Student Member,~IEEE,}
Yunfei Zhu~\orcidlink{0009-0008-1313-2789},~\IEEEmembership{Student Member,~IEEE,}
Jingheng Yan~\orcidlink{0009-0002-8844-402X},~\IEEEmembership{Student Member,~IEEE,}
Lei Ren~\orcidlink{0000-0001-6346-6930},~\IEEEmembership{Senior Member,~IEEE,}
and Laurence T. Yang~\orcidlink{0000-0002-7986-4244},~\IEEEmembership{Fellow,~IEEE.}

}

\thanks{
    {	
The research is supported by The NSFC (National Science Foundation of China) project No.62225302, 623B2014,  62173023. (Corresponding authors: Lei Ren.)

Haiteng Wang, Yikang Li, and Jingheng Yan are with the School of Automation Science and Electrical Engineering, Beihang University, Beijing 100191, China (email: wanghaiteng@buaa.edu.cn, liyikang@buaa.edu.cn, yjh967@buaa.edu.cn).

Yunfei Zhu is with the School of Software, Beihang University, Beijing 100191, China (email: zhuyunfei@buaa.edu.cn).

Lei Ren is with the School of Automation Science and Electrical Engineering, Beihang University, Beijing 100191, China, also with the Hangzhou International Innovation Institute, Beihang University, Hangzhou 311115, China, and also with the State Key Laboratory of Intelligent Manufacturing System Technology, Beijing 100854, China (e-mail: renlei@buaa.edu.cn).

Laurence T. Yang is with the School of Computer and Artificial Intelligence, Zhengzhou University, Zhengzhou 450001, China, and also with the Department of Computer Science, St. Francis Xavier University, Antigonish, NS B2G 2W5, Canada (e-mail: ltyang@ieee.org).
    }
}
}

\maketitle

\begin{abstract}
Accurate analysis of industrial time-series big data is critical for the Prognostics and Health Management (PHM) of industrial equipment. While recent advancements in Large Language Models (LLMs) have shown promise in time-series analysis, existing methods typically focus on single-modality adaptations, failing to exploit the complementary nature of temporal signals, frequency-domain visual representations, and textual knowledge information. In this paper, we propose TS-MLLM, a unified multi-modal large language model framework designed to jointly model temporal signals, frequency-domain images, and textual domain knowledge. Specifically, we first develop an Industrial time-series Patch Modeling branch to capture long-range temporal dynamics. To integrate cross-modal priors, we introduce a Spectrum-aware Vision-Language Model Adaptation (SVLMA) mechanism that enables the model to internalize frequency-domain patterns and semantic context. Furthermore, a Temporal-centric Multi-modal Attention  Fusion (TMAF) mechanism  is designed to actively retrieve relevant visual and textual cues using temporal features as queries, ensuring deep cross-modal alignment. Extensive experiments on multiple industrial benchmarks demonstrate that TS-MLLM significantly outperforms state-of-the-art methods, particularly in few-shot and complex scenarios. The results validate our framework's superior robustness, efficiency, and generalization capabilities for industrial time-series prediction.
\end{abstract} 


\section{Introduction}

Industrial time-series big data is the cornerstone of Prognostics and Health Management (PHM), essential for ensuring equipment reliability\cite{feng2022spatial}. Rather than simple 1D signals, industrial time-series data involves the integration of multi-modal derived information, such as raw signals, frequency spectrum images, and textual semantic context. The analysis of big data from industrial sensors enables the precise capture of dynamic state variations and the proactive identification of latent failures, thereby significantly enhancing equipment reliability\cite{hussain2024big, liang2023survey}.

Deep learning methods, including RNNs\cite{zhang2023data,zhou2021novel}, CNNs\cite{jin2023adaptive}, Transformers\cite{ren2023dynamic}, and Diffusion models\cite{ren2024mts,wang2025meta}, have been widely explored for PHM tasks. These methods leverage advanced techniques such as frequency decomposition, signal feature extraction, and multi-scale learning to model complex temporal dynamics. However, despite their high accuracy, these models suffer from limited generalization, especially in few-shot and zero-shot scenarios.

Pre-trained Large Language Models (LLMs)\cite{hurst2024gpt,yang2025qwen3,guo2025deepseek} have emerged as a promising solution. Benefiting from training on vast corpora, these models exhibit robust generalization capabilities, making them helpful for complex time-series analysis. This paradigm can be further divided into two strategies: encoder-based and prompt-based methods. In the former, LLMs are treated as high-capacity feature encoders to directly process raw or patched time-series data \cite{yu2023harnessing, zhou2023one}, effectively utilizing the model's pre-trained attention mechanisms to extract universal temporal representations. The latter strategy, known as prompt-based modeling (e.g., PromptCast \cite{xue2023promptcast}), reformulates continuous data into discrete textual descriptions, attempting to activate the LLM's inherent zero-shot reasoning through language prompts. 

Building upon these foundations, recent research has shifted towards leveraging multi-modal large language models (MLLMs) to bridge the gap between time-series data and high-level semantic fusion. This evolution has led to two distinct augmented paradigms: text-augmented and vision-augmented modeling. Text-augmented frameworks, such as Time-LLM \cite{jin2024time}, align temporal patches with textual domain knowledge, enabling the model to incorporate expert priors into its prognostic reasoning. Simultaneously, vision-augmented approaches (e.g., Vision-TS \cite{chen2025visionts}, DiagLLM \cite{wang2025diagllm}) transform 1D sequences into 2D spectrograms or recurrence plots, tapping into the superior feature extraction power of visual-language encoders to capture morphological fault signatures.

Although recent progress has been made in incorporating textual and visual information into modeling, the joint modeling of industrial time-series data with these modalities remains scarcely investigated. Current studies are typically limited to single-modality adaptations, failing to exploit the intrinsic complementarity between temporal, visual, and textual information. Specifically, time-series models capture fine-grained dynamics but miss global morphological patterns, while vision-based models grasp structural signatures but lose temporal resolution. Furthermore, the representation misalignment between continuous signals and discrete tokens remains a significant hurdle. Consequently, there is an urgent need for a unified multi-modal framework that can synergize these diverse information to enhance robustness and generalization in complex industrial environments.

To bridge this gap, we propose TS-MLLM, a unified multi-modal framework that exploits MLLMs to jointly model temporal, visual, and textual information. In our paradigm, text conveys domain knowledge, visual features (derived from frequency-domain representations) capture spectral patterns, and time-series encoding represents continuous temporal dynamics. Specifically, TS-MLLM first employs an industrial time-series patch modeling branch to adaptively capture long-range temporal evolutions through patch-based Transformer blocks. To leverage cross-modal priors, a Spectrum-aware Vision-Language Model Adaptation (SVLMA) module is introduced to align frequency-domain spectrum images with textual domain knowledge within a pre-trained LLM space. Finally, a Temporal-centric Multi-modal Attention Fusion (TMAF) mechanism utilizes temporal features as queries to actively retrieve and integrate supportive cues from the MLLM via a multi-modal attention mechanism. This deep alignment ensures robust performance in complex industrial tasks, such as remaining useful life (RUL) estimation. The main contributions of this work are:

1) We propose TS-MLLM, a multi-modal large language model-based framework to jointly model frequency-domain images, textual knowledge, and temporal signals, effectively exploiting cross-modal complementarity to improve the generalization of industrial time-series prediction.

2) We propose spectrum-aware vision-language model adaptation that jointly encodes spectral and semantic features through dual-branch learning, enabling vision-language models to internalize frequency-domain dynamics for enhanced multi-modal reasoning.

3) We develop a temporal-centric multi-modal attention fusion mechanism that treats each temporal feature as a query to retrieve the most relevant visual and textual details via cross-attention, enabling the model to actively integrate complementary multi-modal cues and enhance temporal understanding.

4) Extensive experiments on multiple industrial benchmark datasets demonstrate that TS-MLLM consistently outperforms state-of-the-art baselines in both few-shot and complex scenarios, validating its robustness, efficiency, and generalization.

The remainder of this paper is organized as follows. Section II reviews related studies on deep learning and multi-modal large language models for time-series analysis. Section III presents the architecture and key components of the proposed framework in detail. Experimental settings, results, and corresponding discussions are provided in Section IV. Finally, Section V summarizes the main findings and outlines future research directions.

\section{Related works}
\subsection{Deep Learning for Industrial Time-series}
Leveraging multi-sensor data, deep learning-based PHM paradigms enable essential functions such as anomaly detection \cite{ding2024alad,zhong2025patchad}, intelligent fault diagnosis\cite{jin2022time,lian2024universal}, and RUL prediction \cite{ren2024dlformer, zhang2023data, xu2023multi}. Deep learning has become the dominant paradigm in PHM due to its powerful feature representation capabilities. RNNs and their variants (e.g., LSTM, GRU) are widely used to model temporal dependencies in sensor signals, enabling effective RUL prediction and anomaly detection. For instance, attention-augmented frameworks like BiLSTM-AT \cite{shah2024novel} were introduced to prioritize critical time steps, though their rigid attention mechanisms can limit adaptability across diverse operating conditions. CNNs extract spatially localized features and have been applied to spectrograms or vibration maps to capture degradation patterns. Hybrid architectures, such as CNN-LSTM \cite{ouyang2024combined} and CNN-BiLSTM-AT \cite{yuan2025lithium}, balance local feature extraction with global temporal modeling to enhance representational expressiveness. Recently, Transformer-based models\cite{ren2023dynamic} leverage parallel self-attention to capture multi-scale dependencies efficiently and handling complex temporal correlations. 

However, these methods remain constrained by limited generalization across equipment types and working conditions.

\subsection{Multi-modal Large Language Models for Time-series }

To overcome the limitations of uni-modal approaches and the risks of representation misalignment, researchers have increasingly turned to auxiliary modalities—such as textual descriptions and visual representations—to augment time-series analysis. These methods leverage the robust generalization and transfer capabilities of MLLMs. Current LLM-based approaches can be broadly categorized into three distinct paradigms: time-series based, text-based, and vision-based models.

Time-series based foundation models: The first paradigm focuses on developing foundation models trained from scratch on large-scale, cross-domain time-series datasets\cite{ansari2024chronos, yuan2024unist,rasul2023lag,goswami2024moment,wang2024timexer,das2024decoder}. Representative frameworks, such as those proposed by Ansari et al.\cite{ansari2024chronos} and Das et al.\cite{das2024decoder}, aim to learn universal temporal representations directly from continuous data. However, unlike natural language or computer vision domains, time-series data exhibit high heterogeneity regarding sequence length, sampling rates, variable dimensions, and semantic meanings.  However, the intrinsic diversity of time-series data presents ongoing challenges for pre-trained models to achieve consistent performance across diverse downstream industrial tasks.

Text-augmented models: The second category leverages the reasoning and semantic understanding capabilities of LLMs\cite{jin2024time, lan2025gem, chen2025domain}. These approaches integrate Natural Language Processing (NLP) techniques into time-series analysis to extract semantic context from associated text. For instance, Time-LLM\cite{jin2024time} aligns temporal signals with textual prompts and descriptions, enabling the model to utilize linguistic prior knowledge to assist in forecasting. Similarly, approaches like DK-ILM \cite{lan2025gem}are tailored for RUL prediction. Despite these semantic gains, textual descriptions alone often fail to capture the subtle, high-frequency physical fluctuations inherent in raw industrial signals.

Vision-augmented models: The third paradigm utilizes Vision-Language Models (VLMs) by treating time-series data as visual signals. These methods convert one-dimensional temporal sequences into two-dimensional representations (e.g., time-frequency spectrograms) to exploit the powerful feature extraction capabilities of visual encoders. Vision-TS\cite{chen2025visionts}, for example, processes visualized signal states to capture morphological patterns that are difficult to detect in raw numerical data. In the domain of fault diagnosis, DiagLLM\cite{wang2025diagllm} integrates envelope spectrum images with expert knowledge. By employing VLMs to align visual fault signatures with linguistic reasoning, such models achieve a deeper understanding of equipment states, facilitating more accurate diagnosis and prediction.

\begin{figure*}[t]
	\centering
	\includegraphics[width=2\columnwidth]{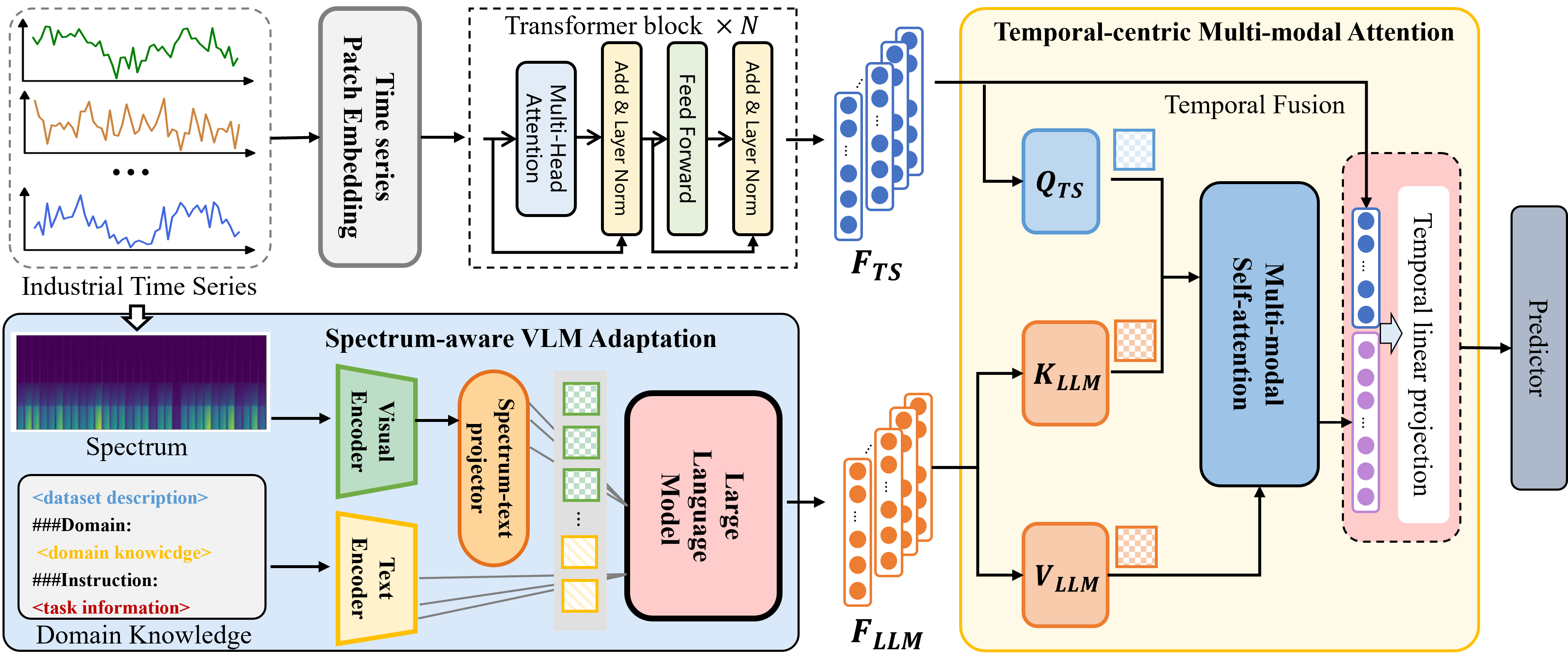}
	\caption{The overall architecture of the proposed TS-MLLM. It consists of three primary modules: (1) An Industrial time-series Modeling Branch for capturing long-range temporal dependencies; (2) SVLMA aligns frequency-domain representation with textual domain knowledge via MLLM; and (3) TMAF mechanism utilizes temporal features as queries to actively retrieve and integrate multi-modal cues for final prediction.}
	\label{fig:architecture}
\end{figure*}

These methods demonstrate the potential of transferring knowledge from general LLMs to industrial PHM tasks. Nonetheless, most existing models rely on single modalities, overlooking the complementary nature of temporal, visual, and textual information, which limits their robustness and adaptability in real-world industrial environments.

\section{Methodology}

\subsection{Model Structure}
The TS-MLLM architecture comprises three primary components: an industrial time-series patch modeling branch, a spectrum-aware vision-language model adaptation module, and a temporal-centric multi-modal attention fusion mechanism.

\paragraph{Industrial Time-series Patch Modeling}
Input industrial signals are first converted into high-dimensional representations via patch embedding. These features are processed by $N$ Transformer blocks, utilizing multi-head attention and feed-forward networks to capture non-linear dependencies and long-range temporal evolutions. The resulting feature vectors, $\mathbf{F}_{TS}$, serve as the primary queries for multi-modal fusion.

\paragraph{Spectrum-aware Vision-Language Model Adaptation}
To leverage cross-modal priors, raw signals are transformed into spectrum images and integrated with textual domain knowledge. Following feature extraction by dedicated encoders, a spectrum-text projector aligns these features into a unified embedding space. A pre-trained LLM then processes these tokens to internalize frequency-domain dynamics and semantic information, outputting a guided representation $\mathbf{F}_{LLM}$.

\paragraph{Temporal-centric Multi-modal Attention Fusion}
Finally, a TMAF mechanism is proposed to achieve deep alignment between modalities. 
This mechanism employs the temporal branch output as the query vector $\mathbf{Q}_{TS}$. 
It actively retrieves relevant information from the MLLM-derived features, which are projected into key and value vectors, $\mathbf{K}_{LLM}$ and $\mathbf{V}_{LLM}$, facilitating a selective search for supportive visual cues. 
This process generates a context vector $\mathbf{F}_{attn}$ weighted by temporal relevance, which is subsequently concatenated with the original signals and fused via a temporal linear projection. 
After which a predictor generates the final results, such as RUL estimation or fault diagnosis.

\subsection{Time-series Patch Modeling}

The primary challenge in regression modeling for industrial multivariate time-series lies in the inherent complexity of capturing long-range dependencies and dynamic evolutions within high-dimensional sensor data. Unlike simple sequential data, industrial signals are characterized by intricate temporal dynamics, non-stationarity, and complex multivariate correlations, where the critical information regarding the system's state is distributed across extended historical windows rather than being observable at isolated time steps.

Traditional pointwise Transformer architectures treat individual time steps as semantic tokens. However, this approach faces two fundamental limitations in the current context. First, the computational cost grows quadratically with sequence length, which limits the feasible input size and prevents the model from capturing the long-term dependencies required for precise regression. Second, they lack the capacity to aggregate local contextual information, leading to fragmented representations that fail to capture the coherent continuous state evolution of the complex industrial system.

To address these limitations and capture both local semantic patterns and long-range dependencies, we propose a Patch-based Temporal Modeling framework. Instead of operating on single time steps, we implement a segmentation strategy that groups adjacent time points into meaningful sub-series units.

We define the input multivariate time-series as $X \in \mathbb{R}^{L \times M}$, where $L$ denotes the input sequence length and $M$ represents the number of sensor channels. To mitigate boundary effects and ensure complete feature extraction at the sequence edges, we first apply a replication padding operation to the temporal dimension. Subsequently, by deploying a sliding window with patch length $P$ and a unit stride $S=1$, we unfold the series into a tensor of shape $\mathbb{R}^{N \times (P \cdot M)}$. This dense overlapping strategy ensures continuity between adjacent tokens, where the total number of patches is derived as $N = \frac{L-P}{S} + 2$.

By treating patches as the fundamental modeling units, we naturally capture the local context that pointwise approaches overlook. To transform these raw sub-series into learnable feature vectors, we employ a linear projection $W_{emb}$. Additionally, to preserve the temporal order of the sequence, learnable positional encodings $E_{pos}$ are added. We segment the input time-series into a sequence of flattened patches $X_p \in \mathbb{R}^{N \times (P \cdot M)}$. The input embedding is then obtained via:
\begin{equation}
    Z_0 = X_p W_{emb} + E_{pos}
\end{equation}
This sequence $Z_0$ serves as the input to the Transformer encoder. By operating on these patch-level representations, the self-attention mechanism efficiently models global dependencies across the entire look-back window:
\begin{equation}
    \mathbf{F}_{TS} = \text{Encoder}(Z_0)
\end{equation}
The output of this branch, $\mathbf{F}_{TS} \in \mathbb{R}^{N \times D_{model}}$, provides a comprehensive representation of the time-series evolution, serving as the temporal alignment reference for the subsequent multi-modal integration. Crucially, during the initial training phase, this branch is optimized independently. This isolation ensures that the model extracts intrinsic degradation dynamics exclusively from temporal data, establishing a robust unimodal representation prior to the integration of heterogeneous multi-modal information.

\subsection{Spectrum-aware Vision-Language Model Adaptation (SVLMA)}

While temporal modeling captures the evolution of degradation, industrial faults often manifest as distinct, texture-like patterns in the frequency domain that are obfuscated in the time domain. Furthermore, pure data-driven approaches often neglect the rich semantic context provided by domain experts, such as operating conditions or asset specifications.

To bridge this gap, we introduce the SVLMA module. This architecture is designed to construct a unified semantic manifold where spectral visual patterns and expert textual knowledge are aligned via a LLM, serving as a global context generator for the industrial process.

As illustrated in the Figure, this module is designed to construct a unified semantic manifold through three functional components: the Time-Frequency Transformation ($\text{TFT}$) module, the Domain Knowledge Embedding (DKE) module and the vision-language model adaptation (VLMA) module . Accordingly, we formulate the internal inference flow as follows:
\begin{align}
    \mathbf{F}_{spec} &= \text{TFT}(X) \\
    \mathbf{F}_{text} &= \text{DKE}(X_{text}) \\
    \mathbf{F}_{LLM} &= \text{VLMA}(\mathbf{F}_{spec}, \mathbf{F}_{text})
\end{align}
This formulation explicitly delineates the coordination between these stages: (1) $\text{TFT}$ transforms the raw time-series $X$ into spectral visual representations $\mathbf{F}_{spec}$, (2) $\text{DKE}$ encodes the expert domain knowledge $X_{text}$ into structured textual embeddings $\mathbf{F}_{text}$, and (3) $\text{VLMA}$ aligns these multi-modal features to generate the global semantic context $\mathbf{F}_{LLM}$.

\subsubsection{Time-Frequency Transformation}

To translate 1D temporal dynamics into the 2D visual domain required by the VLMA, standard approaches typically rely on fixed operators like the Short-Time Fourier Transform (STFT). However, these methods suffer from rigid, predefined basis functions that are often suboptimal for capturing the non-stationary and transient fault signatures inherent in complex industrial data.

To overcome this limitation, we introduce a Multi-view Time-Frequency Transformation layer. Instead of a single view, we construct a composite 3-channel tensor $I_{spec}$ designed to integrate diverse physical signal characteristics. We initially employ Recurrence Plots (RP) to encode non-linear system dynamics. This operator maps 1D temporal trajectories into 2D spatial textures, effectively highlighting recurring system states often missed by linear methods. Given the reconstructed phase space vectors $v_i$ and $v_j$, the recurrence matrix $R$ is derived as:
\begin{equation}
    R_{i,j} = \Theta(\epsilon - \|v_i - v_j\|)
\end{equation}
where $\epsilon$ denotes the threshold distance and $\Theta$ is the Heaviside function.

Complementing this topological view, the STFT is applied to extract stationary spectral features. This step directly converts time-domain signals into frequency-domain heatmaps, revealing periodic patterns and energy distributions across frequency bands. The spectrogram $S_{stft}$ is computed by sliding a window function $w(\cdot)$ over the signal $x(t)$:
\begin{equation}
    S_{stft}(t, \omega) = \int_{-\infty}^{\infty} x(\tau) w(\tau - t) e^{-j\omega\tau} d\tau
\end{equation}
To resolve transient impulses where STFT falls short due to fixed windowing, we incorporate the Continuous Wavelet Transform (CWT). This multi-scale analysis precisely localizes short-term variations in the time-frequency plane. The wavelet coefficients $S_{cwt}$ at scale $a$ and translation $b$ are determined by the convolution with a mother wavelet $\psi$:
\begin{equation}
    S_{cwt}(a, b) = \frac{1}{\sqrt{a}} \int_{-\infty}^{\infty} x(t) \psi^*\left(\frac{t-b}{a}\right) dt
\end{equation}
These three complementary representations are synthesized into a unified 'RGB-like' tensor via channel-wise concatenation:
\begin{equation}
    \mathbf{F}_{spec} = \text{Concat}(R, S_{stft}, S_{cwt})
\end{equation}
This design enables the vision encoder to simultaneously process topological, spectral, and time-scale features. By converting diverse signal characteristics into a standard image format, we fully leverage the pattern recognition power of the pre-trained model.

\subsubsection{Domain Knowledge-guided Text Embedding}

While time-series and frequency data are informative, they inherently lack explicit high-level context regarding the operational environment. To bridge this gap, we implement a knowledge injection mechanism that synthesizes semantic domain knowledge into structured textual prompts, denoted as $X_{text}$.

Once the text is constructed, it undergoes a rigorous tokenization process to convert the raw strings into a machine-readable format. The text $X_{text}$ is first normalized and segmented into sub-word units using algorithms such as Byte-Pair Encoding (BPE), which allows for the robust handling of rare, domain-specific terminology. These sub-word tokens are then mapped to a sequence of numerical indices based on a fixed vocabulary, with the sequence being either padded or truncated to match a pre-set maximum length $L$. Formally, this mapping from the raw text to the integer index sequence $S$ can be expressed as:
\begin{equation}
    S = \text{Tokenizer}(X_{text}) = \{w_1, w_2, \dots, w_L\} \in \mathbb{R}^L
\end{equation}
where $w_i$ represents the unique index of the $i$-th token in the vocabulary.

Following the conversion of text into discrete indices, we project the sequence into a continuous latent space to preserve semantic information. This transformation is achieved by utilizing a learnable embedding matrix $W_C$, which converts scalar indices into dense vectors. To retain the temporal structure of the sequence, we inject positional information via $W_{pos}$. Consequently, the resulting textual embedding $\mathbf{F}_{text}$ is defined as:
\begin{equation}
    \mathbf{F}_{text} = \text{Embed}(S) + W_{pos}
\end{equation}
where $\text{Embed}(\cdot)$ denotes the embedding lookup operation parameterized by $W_C \in \mathbb{R}^{V \times d}$, and $W_{pos}$ serves to encode the positional context within the sequence.

\subsubsection{Vision-Language Model Adaptation}

In the preceding phases, we successfully transformed raw time-series data into spectral images $\mathbf{F}_{spec}$ and constructed domain-specific textual prompts. However, these two modalities reside in isolated feature spaces. To synthesize them, we propose the Vision-Language Model Adaptation module. This architecture is specifically designed to bridge the gap between low-level signal processing and high-level semantic reasoning.

We initiate the encoding process by handling the spectral component. A pre-trained Masked Autoencoder serves as the visual backbone $V(\cdot)$. This model functions as a sophisticated feature extractor. It ingests the high-dimensional spectral images and compresses the frequency-domain information into a compact visual embedding $h_{vis} \in \mathbb{R}^{D_{vis}}$. This vector effectively summarizes the holistic health signature of the equipment:
\begin{equation}
    h_{vis} = V(\mathbf{F}_{spec})
\end{equation}
Merely extracting visual features is insufficient. We must integrate these features with the semantic capabilities of the Qwen Large Language Model. A fundamental challenge arises from the dimensional mismatch between the visual latent space and the LLM's token embedding space. We resolve this issue through a learnable cross-modal projector $\phi$. This module linearly maps the visual embedding $h_{vis}$ into the semantic dimension $\mathbb{R}^{D_{llm}}$. The projection aligns the physical signal with the linguistic manifold.

The resulting visual token $e_{vis} = \phi(h_{vis})$ acts as a continuous visual prefix. It translates abstract physical signals into a format that the LLM can interpret. We then actuate the fusion process. The projected visual token $e_{vis}$ is prepended to the text token embeddings $F_{text}$. This operation constructs a composite input sequence for the multi-modal reasoning engine. The LLM is forced to attend to both physical evidence and semantic context simultaneously:
\begin{equation}
    \mathbf{F}_{LLM} = \text{LLM}(\text{Concat}[e_{vis}, \mathbf{F}_{text}])
\end{equation}
The LLM processes this unified sequence to generate a deep semantic representation $\mathbf{F}_{LLM}$. This output constitutes the LLM-based feature modeling of the raw time-series data. It effectively solves the limitation of traditional single-modality analysis. By fusing the observational reality of spectral data with the inferential logic of textual knowledge, we obtain a comprehensive understanding of the equipment state, providing a robust and noise-resilient foundation for subsequent fault diagnosis tasks.

\subsection{Temporal-centric Multi-modal Attention Fusion (TMAF)}

Given the distinct feature representations obtained from the temporal backbone and the vision-language branch, the final challenge lies in their effective integration. Naive fusion strategies, such as global average pooling, are suboptimal. These methods often discard the temporal resolution required to pinpoint the exact onset of degradation.

To address this limitation, we introduce the TMAF mechanism. Our design philosophy prioritizes the patch-wise time-series features as the primary information carrier. This approach preserves the structural integrity of the degradation trajectory. Simultaneously, the multi-modal semantic features function as a global reference to modulate the temporal predictions.

We formulate this fusion as an asymmetric Query-Key-Value attention operation across distinct domains. The temporal features $\mathbf{F}_{TS} \in \mathbb{R}^{N \times D_{model}}$ retain the sequence length $N$. These features generate the Query matrix $\mathbf{Q}_{TS}$ via a linear transformation $W_Q$. In parallel, the integrated semantic representation $\mathbf{F}_{LLM}$ encapsulates the global health context. This vector is projected via matrices $W_K$ and $W_V$ to form the Key ($\mathbf{K}_{LLM}$) and Value ($\mathbf{V}_{LLM}$) matrices. Since $\mathbf{F}_{LLM}$ represents the entire window globally, we broadcast it along the temporal dimension. This alignment matches the sequence length of the Query. Mathematically, the projections are defined as:
\begin{align}
    \mathbf{Q}_{TS} &= \mathbf{F}_{TS}W_Q, \\
    \mathbf{K}_{LLM} &= \mathbf{F}_{LLM}W_K, \\
    \mathbf{V}_{LLM} &= \mathbf{F}_{LLM}W_V
\end{align}
where $\mathbf{Q}_{TS} \in \mathbb{R}^{N \times D_k}$ denotes the query matrix corresponding to the temporal sequence of length $N$. The key and value matrices, $\mathbf{K}_{LLM}$ and $\mathbf{V}_{LLM}$, are derived from the global context feature $\mathbf{F}_{LLM}$. To ensure dimensional compatibility, the global context is replicated along the temporal axis, resulting in effective dimensions of $\mathbf{K}_{LLM}, \mathbf{V}_{LLM} \in \mathbb{R}^{N \times D_k}$. This asymmetric configuration enables each temporal segment to retrieve relevant auxiliary information, aligning local signal fluctuations with the global spectral and semantic context. 

The interaction between the temporal trajectory and the context from multiple sources is quantified using the scaled dot product. This calculation yields an attention map $A$. The map measures the relevance of the global context to each local temporal segment:
\begin{equation}
    A = \text{Softmax}\left(\frac{\mathbf{Q}_{TS}\mathbf{K}_{LLM}^T}{\sqrt{D_k}}\right)
\end{equation}
The attention score $A_i$ for the $i$-th time step reflects the semantic alignment between local behavior and the global health state. For instance, if the LLM identifies a critical spectral signature, the mechanism emphasizes specific time steps with corresponding anomalous fluctuations. This process filters out irrelevant noise.

The retrieved context information is computed as $\mathbf{F}_{attn} = A\mathbf{V}_{LLM}$. Subsequently, this context is fused back into the temporal stream. To ensure gradient stability and preserve pre-trained temporal knowledge, we employ a residual connection followed by a final linear projection $W_{out}$:
\begin{equation}
    \mathbf{F}_{fused} = \text{Concat}[\mathbf{F}_{TS}, \mathbf{F}_{attn}]W_{out}
\end{equation}
The context enriched sequence $\mathbf{F}_{fused}$ is subsequently processed by the regression head to yield the final industrial forecast. This output represents a rigorous synthesis of observed signal patterns and inferential semantic knowledge. By structurally prioritizing temporal dynamics while integrating global context, the framework effectively exploits the complementarity between distinct data domains. This architecture mitigates the interference of noise and enhances the reasoning capability of the model. Consequently, this architecture establishes a resilient foundation for generalization across complex industrial scenarios, aligning with the core design objectives of our unified multi-modal framework.

\section{Experimental Results}
\subsection{Experimental Setup}

\subsubsection{Dataset}
The Commercial Modular Aero-Propulsion System Simulation (C-MAPSS)\cite{saxena2008damage} dataset, released by NASA in 2008, serves as a benchmark for characterizing the degradation process of turbofan engines. This dataset comprises 21 multi-sensor variables, including temporal data, pressure, and rotational speeds, which collectively monitor the operational health of the engines. However, certain sensors (specifically indices 1, 5, 6, 10, 16, 18, and 19) remain constant throughout the lifecycle and lack correlation with the RUL. To eliminate redundant information and enhance computational efficiency, these seven inactive sensors were excluded, resulting in 14 effective sensory features for subsequent analysis.

As listed in Table \ref{tab:1}, C-MAPSS consists of four distinct sub-datasets (FD001–FD004) categorized by environmental complexity. While FD001 and FD003 involve a single operating condition and a unique fault mode, FD002 and FD004 incorporate six operating conditions and up to two fault types. The multi-condition nature of the latter subsets provides a more authentic representation of the complexities inherent in real-world industrial systems.

\begin{center}
\begin{table}[htpb]
	\centering
	\caption{Description of C-MAPSS Datasets }
	\label{tab:1}       
	\begin{tabular}{lllll}
	\toprule 
		Sub-dataset	& FD001	& FD002	& FD003	& FD004  \\
		\midrule 
		Training engines	&100	&260	&100	&249 \\
		Testing engines	&100	&259	&100	&248 \\
		Operation conditions	&1	&6	&1	&6 \\
		Fault modes	&1	&1	&2	&2 \\

	\bottomrule	
	\end{tabular}
\end{table}
\end{center}

\subsubsection{Hyper-parameters}
\begin{center}
\begin{table}[htbp]
  \centering
  \caption{Hyperparameter Settings for Each Dataset}
  \label{tab:hyperparameters}
  \resizebox{\linewidth}{!}{
    \begin{tabular}{llcccc} 
    \toprule
    Category & Parameter & FD001 & FD002 & FD003 & FD004 \\
    \midrule
    
    \multirow{3}{*}{Optimization} 
      & Batch Size & 128 & 128 & 128 & 128 \\
      & Learning Rate & 0.002 & 0.002 & 0.002 & 0.002 \\
      & Total Epochs & 30 & 30 & 30 & 30 \\
    \midrule
    
    \multirow{2}{*}{Data Input} 
      & Window Size ($L$) & 40 & 40 & 40 & 40 \\
      & Input Channels ($M$) & 14 & 14 & 14 & 14 \\
    \midrule
    
    \multirow{4}{*}{Temporal Branch} 
      & Patch Size ($P$) & 4 & 4 & 4 & 4 \\
      & Patch Stride ($S$) & 1 & 1 & 1 & 1 \\
      & Model Dim & 64 & 64 & 64 & 64 \\
      & Attention Heads & 1 & 1 & 1 & 1 \\
    \midrule
    
    \multirow{4}{*}{\shortstack[l]{Spectrum-aware\\VLMA}} 
      & Image Resolution & 114*114 & 114*114 & 114*114 & 114*114 \\
      & Visual Embed & 128 & 128 & 128 & 128 \\
      & Text Feature Dim & 96 & 96 & 96 & 96 \\
      & Max Token Length & 512 & 512 & 512 & 512 \\
    \midrule
    
    \multirow{4}{*}{TMAF} 
      & Attention Key & 64 & 64 & 64 & 64 \\
      & Attention Output & 32 & 32 & 32 & 32 \\
      & Fusion MLP Units & 512 & 512 & 512 & 512 \\
      & Dropout Rate & 0.5 & 0.5 & 0.5 & 0.5 \\
      
    \bottomrule
    \end{tabular}
  }
\end{table}
\end{center}

\begin{table*}[t]
  \centering
  \caption{Main results on C-MAPSS (FD001--FD004). Lower is better.}
  \label{tab:main_results}
  \setlength{\tabcolsep}{4.2pt}
  \renewcommand{\arraystretch}{1.08}
  \scriptsize
  \resizebox{0.67\linewidth}{!}{
  \begin{tabular}{lcccccccc}
    \toprule
    \multirow{2}{*}{\textbf{Method}} &
    \multicolumn{2}{c}{\textbf{FD001}} &
    \multicolumn{2}{c}{\textbf{FD002}} &
    \multicolumn{2}{c}{\textbf{FD003}} &
    \multicolumn{2}{c}{\textbf{FD004}} \\
    \cmidrule(lr){2-3}\cmidrule(lr){4-5}\cmidrule(lr){6-7}\cmidrule(lr){8-9}
    & \textbf{RMSE} & \textbf{Score} & \textbf{RMSE} & \textbf{Score} & \textbf{RMSE} & \textbf{Score} & \textbf{RMSE} & \textbf{Score} \\
    \midrule
    BiGRU               & 13.71 & 302.27 & 17.34 & 1345.92 & 14.02 & 663.01 & 19.23 & 1605.18 \\
    BiLSTM              & 14.12 & 296.02 & 16.91 & 1316.97 & 15.18 & 906.71 & 23.76 & 4349.31 \\
    MR-LSTM \cite{xu2023multi}  & / & /  & 15.71 & 1434.27 & / & /  & 16.81 & 1785.33 \\
    C-Transformer \cite{ZHOU2023109357}       & 13.79 & 475.46 & 16.11 & 2214.59 & 17.10 & 939.10 & 19.77 & 3237.37 \\
    AMR-Net \cite{Wang2025amr}            & 12.49 & 269.08 & 14.72 & 1085.03 & 13.88 & 372.05 & 16.30 & \textbf{998.57} \\ \hline
    DeeBERT \cite{xin2020deebert}  & 12.75 & 243.64  & 16.32 & 1429.15 & 12.36 & 254.19  & 16.53 & 1671.03 \\
    One Fits All (GPT-2) \cite{zhou2023one}    & 12.61 & 310.98 & 16.62 & 1309.36 & 12.36 & \textbf{236.43} & 18.50 & 2088.39 \\
    One Fits All (Qwen3-0.6B) \cite{zhou2023one,yang2025qwen3} & 13.62 & 259.40 & 21.38 & 4941.72 & 12.45 & 283.98 & 20.69 & 8543.54 \\ 
    \hline
    \textbf{TS-MLLM (ours)} & \textbf{12.45} & \textbf{233.40} & \textbf{14.22} & \textbf{929.81} & \textbf{11.97} & 338.30 & \textbf{15.94} & 1715.11 \\
    \bottomrule
  \end{tabular}
  }
\end{table*}
\subsubsection{Data Process}

\paragraph{Min-Max Normalization} A Min-Max normalization technique is employed to project the raw sensory signals into a unified range of $[0, 1]$. This rescaling procedure is essential for ensuring numerical stability and accelerating gradient convergence during the model training phase. The transformation for a specific feature $x$ is defined as follows:

\begin{equation}
x_{norm} = \frac{x - \min(x)}{\max(x) - \min(x)}
\end{equation}
where $x_{norm}$ denotes the scaled feature value, while $\min(x)$ and $\max(x)$ represent the minimum and maximum boundary values of the respective sensor sequence across the entire dataset. 

\paragraph{Sliding Window Processing} Following the established preprocessing pipeline\cite{ren2024dlformer}, a sliding window of size $S=50$ is implemented to partition the sensory sequences into $T-S+1$ samples, each with a dimension of $50 \times 14$, thereby significantly expanding the available training data. To more accurately reflect the degradation characteristics of turbofan engines, a piecewise linear RUL strategy\cite{ren2024dlformer} is adopted, which caps the target labels at a maximum threshold of $RUL_{max}=125$ during the early stable operating phase.

\subsection{Evaluation Metrics}
We evaluate the proposed model using two commonly used metrics for RUL prediction: the Root Mean Square Error (RMSE) and the score function.

\begin{equation}
\mathrm{RMSE} = \sqrt{\frac{1}{N}\sum_{i=1}^{N}\left(\hat{y}_i - y_i\right)^2},
\end{equation}
where $N$ is the number of samples, and $\hat{y}_i$ and $y_i$ denote the predicted and true RUL of the $i$-th sample, respectively. RMSE reflects the overall prediction accuracy, with a lower value indicating better performance.

The score function is an asymmetric metric that penalizes late predictions more severely than early ones, as overestimation of RUL poses higher risks in practice:

\begin{equation}
\mathrm{Score} =
\begin{cases}
\sum_{i=1}^{N} \left[ \exp\left( -\frac{\hat{y}_i - y_i}{13} \right) - 1 \right], & \hat{y}_i < y_i, \\
\sum_{i=1}^{N} \left[ \exp\left(  \frac{\hat{y}_i - y_i}{10} \right) - 1 \right], & \hat{y}_i \ge y_i.
\end{cases}
\end{equation}


\begin{figure*}[t]
	\centering
	\includegraphics[width=2\columnwidth]{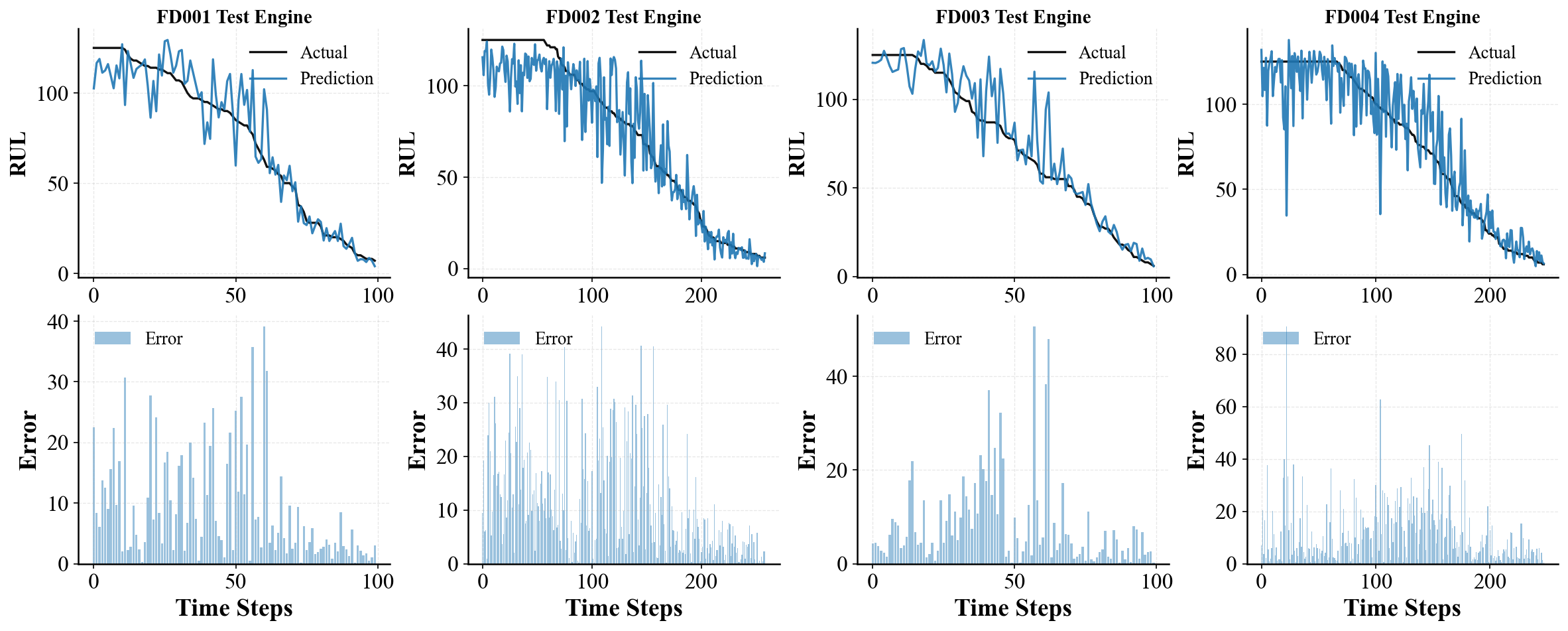}
	\caption{Qualitative RUL prediction on C-MAPSS. Results on representative test engines from FD001 to FD004. For each subset, the top plot shows predicted and ground-truth RUL over time steps, and the bottom plot shows the absolute error over time steps.}
	\label{fig:fd_rul_panels}
\end{figure*}

\begin{figure*}[t]
	\centering
	\includegraphics[width=2\columnwidth]{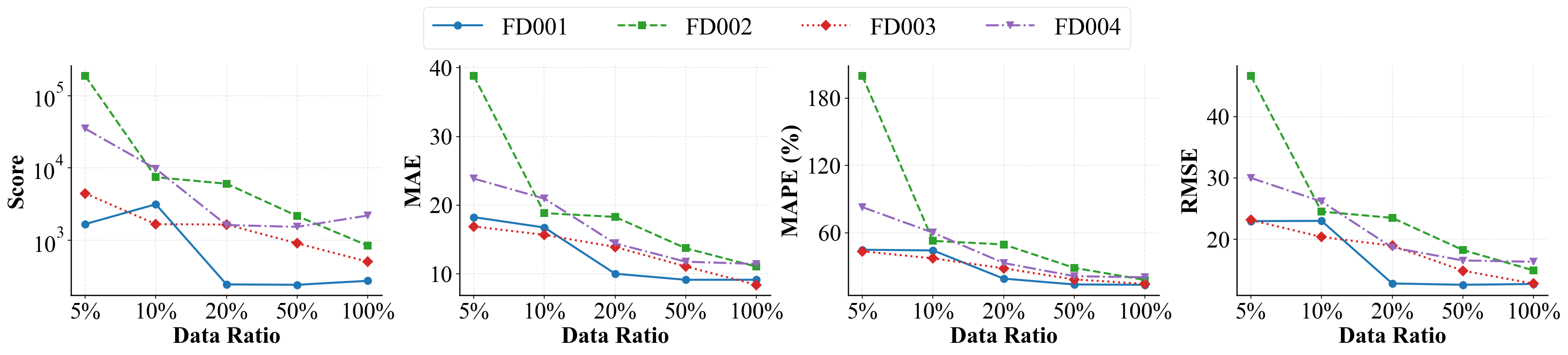}
	\caption{Few-shot performance on C-MAPSS. Results on FD001 to FD004 under different training-data ratios from 5\% to 100\%, evaluated by Score, MAE, MAPE, and RMSE. Lower is better.}
	\label{fig:cmapss_two_scores}
\end{figure*}

\begin{figure}[t]
  \centering
  \includegraphics[width=\columnwidth]{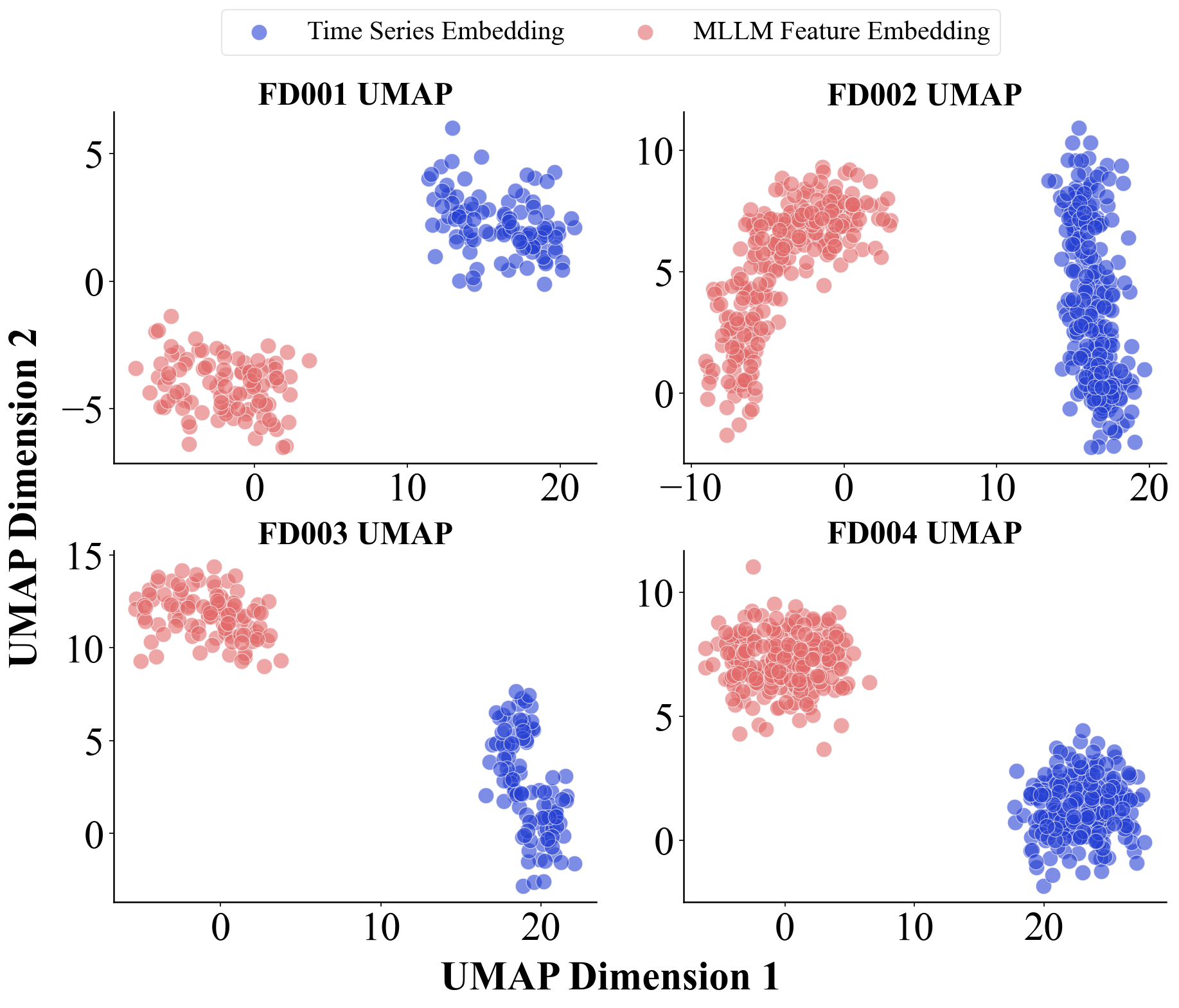}
  \caption{UMAP projections of time-series embeddings and MLLM feature embeddings on FD001--FD004. The two embedding sets are largely separated across subsets.}
  \label{fig:fig3_real_umap_grid}
\end{figure}

\subsection{Main Results}

To systematically assess the RUL regression capability of TS-MLLM, we evaluate on all four subsets FD001 to FD004 and report RMSE and Score, where lower is better.

Table~\ref{tab:main_results} summarizes the comparison with a diverse set of baselines, including recurrent backbones, Transformer-based forecasting, a strong task-specific approach, and recent LLM and VLM inspired methods. TS-MLLM achieves the lowest RMSE on all subsets, obtaining 12.45, 14.22, 11.97, and 15.94 on FD001, FD002, FD003, and FD004, respectively. Relative to the strongest baseline on each subset, this corresponds to RMSE reductions of 0.3\% on FD001, 3.4\% on FD002, 3.2\% on FD003, and 2.2\% on FD004, with an average reduction of about 2.3\%. These consistent gains indicate that TS-MLLM improves regression accuracy under diverse operating conditions and degradation modes.

For Score, TS-MLLM achieves the best performance on FD001 and FD002, yielding 233.40 and 929.81, which correspond to 10.0\% and 14.3\% improvements over the strongest baseline, respectively. On FD003 and FD004, TS-MLLM reports 338.30 and 1715.11 in Score, which are higher than the best competing results. This suggests that the asymmetric penalty in Score is sensitive to a small number of pronounced early or late prediction cases, and therefore may not decrease monotonically with improvements in average error. Despite this, TS-MLLM remains competitive on Score while delivering consistent gains across subsets, indicating a favorable trade-off between overall accuracy and risk-sensitive error patterns under varying operating regimes.

For qualitative verification, we visualize the predicted RUL trajectories together with the ground truth on the test set. Fig.~\ref{fig:fd_rul_panels} shows representative prediction results of TS-MLLM on FD001 to FD004. In each column, the top panel compares the predicted RUL with the ground-truth trajectory over time steps, while the bottom panel depicts the corresponding absolute error over time steps.

Across all subsets, TS-MLLM produces predictions that closely follow the overall degradation trend and remain stable throughout the full life cycle. In the early stage, where the ground-truth RUL is high and degradation signatures are relatively weak, the predicted curves stay well-behaved without apparent drift, indicating robust temporal stability. As the system enters the mid-life stage and the degradation rate increases, TS-MLLM responds promptly to the change in slope and maintains tight tracking, rather than exhibiting delayed reactions or over-smoothing. Near the end of life, the predictions converge toward the ground truth and preserve a consistent decreasing pattern, which is essential for reliable maintenance planning.

The error panels further corroborate these observations. The absolute errors are generally small and do not persist across long intervals, suggesting that TS-MLLM does not suffer from systematic bias accumulation. Occasional error spikes mainly occur around transitional regions where the degradation dynamics change more abruptly, yet the model recovers quickly and returns to accurate tracking. This trend is especially visible on FD002 and FD004, where longer sequences and more complex condition variations still result in coherent trajectories without excessive oscillation. Overall, the qualitative results demonstrate that TS-MLLM yields stable and responsive life-cycle tracking across different C-MAPSS subsets.

\subsection{Few-Shot Learning}
To evaluate the sample efficiency of TS-MLLM under limited-label regimes, we down-sample the training set to 5\%, 10\%, 20\%, and 50\% of the original data, using 100\% training as a full-data reference. Fig.~\ref{fig:cmapss_two_scores} reports the performance trends on the four C-MAPSS subsets as the training ratio varies, measured by Score, MAE, MAPE, and RMSE.

Overall, increasing the training ratio consistently improves performance across subsets and metrics. The most noticeable gains are observed when moving from 5\% to 10\% or 20\%, after which the curves gradually flatten, indicating that TS-MLLM enters a near-saturated regime with a relatively small fraction of labeled data. On FD001 and FD003, the performance at 20\% is already close to that of higher ratios for MAE, MAPE, and RMSE, suggesting strong data efficiency on these subsets. In contrast, FD002 exhibits a much larger gap at 5\% and benefits more substantially from additional supervision, which is consistent with its higher variability in operating conditions. For FD004, the metrics decrease steadily as data increases, while slight fluctuations can be observed between 50\% and 100\% on some metrics.

These results support that TS-MLLM can leverage limited labeled data effectively. The spectral-aware adaptation aligns frequency-domain dynamics with semantic priors, providing useful inductive bias in low-data regimes, and the temporal-centric multi-modal fusion selectively integrates informative cues, which helps stabilize learning when supervision is scarce.

\begin{figure}[t]
  \centering
  \includegraphics[width=\columnwidth]{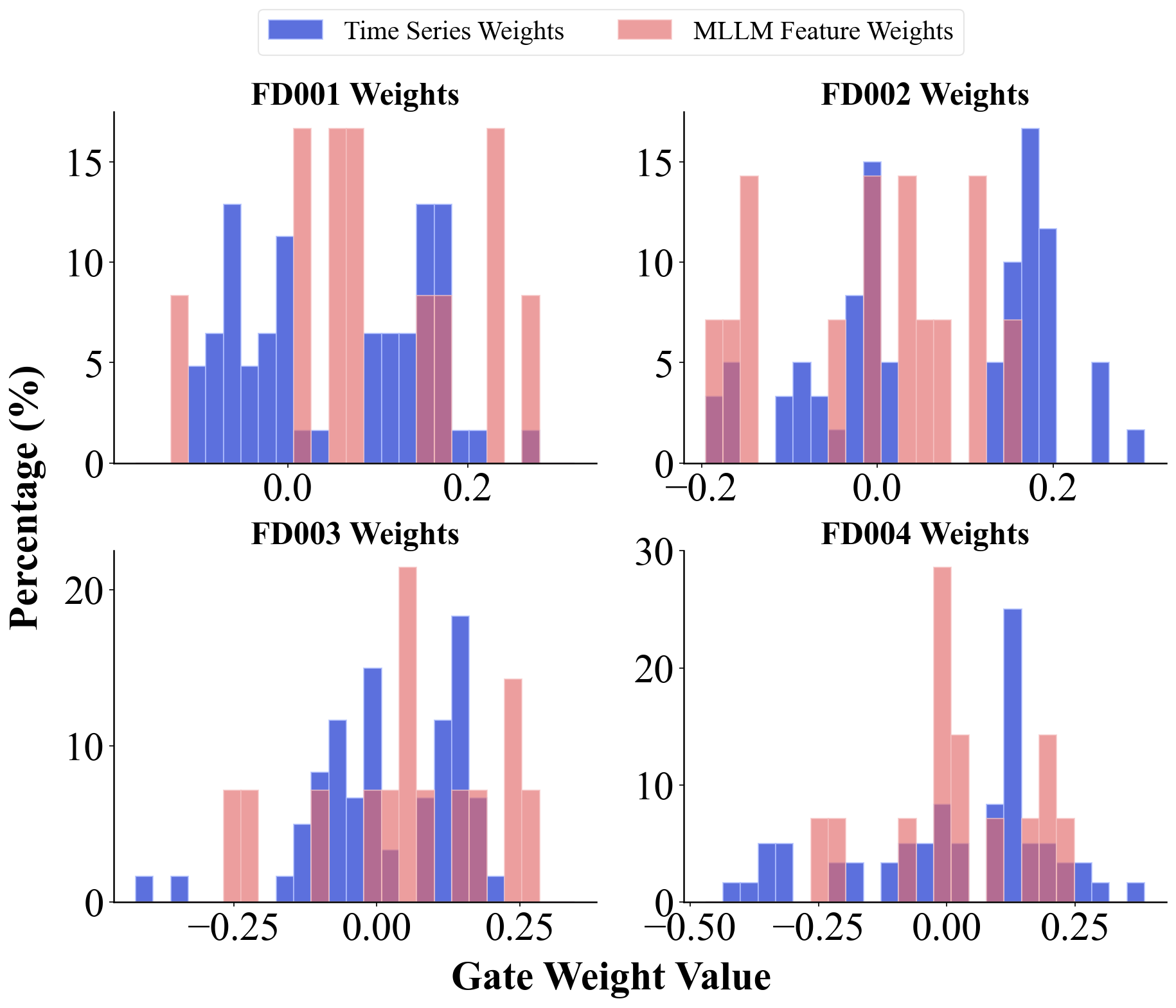}
  \caption{Distributions of fusion-gate weights on FD001--FD004. Weights for time-series and MLLM features span a broad range and vary across subsets.}
  \label{fig:fig3_real_gate_grid}
\end{figure}

\begin{figure}[t]
  \centering
  \includegraphics[width=\columnwidth]{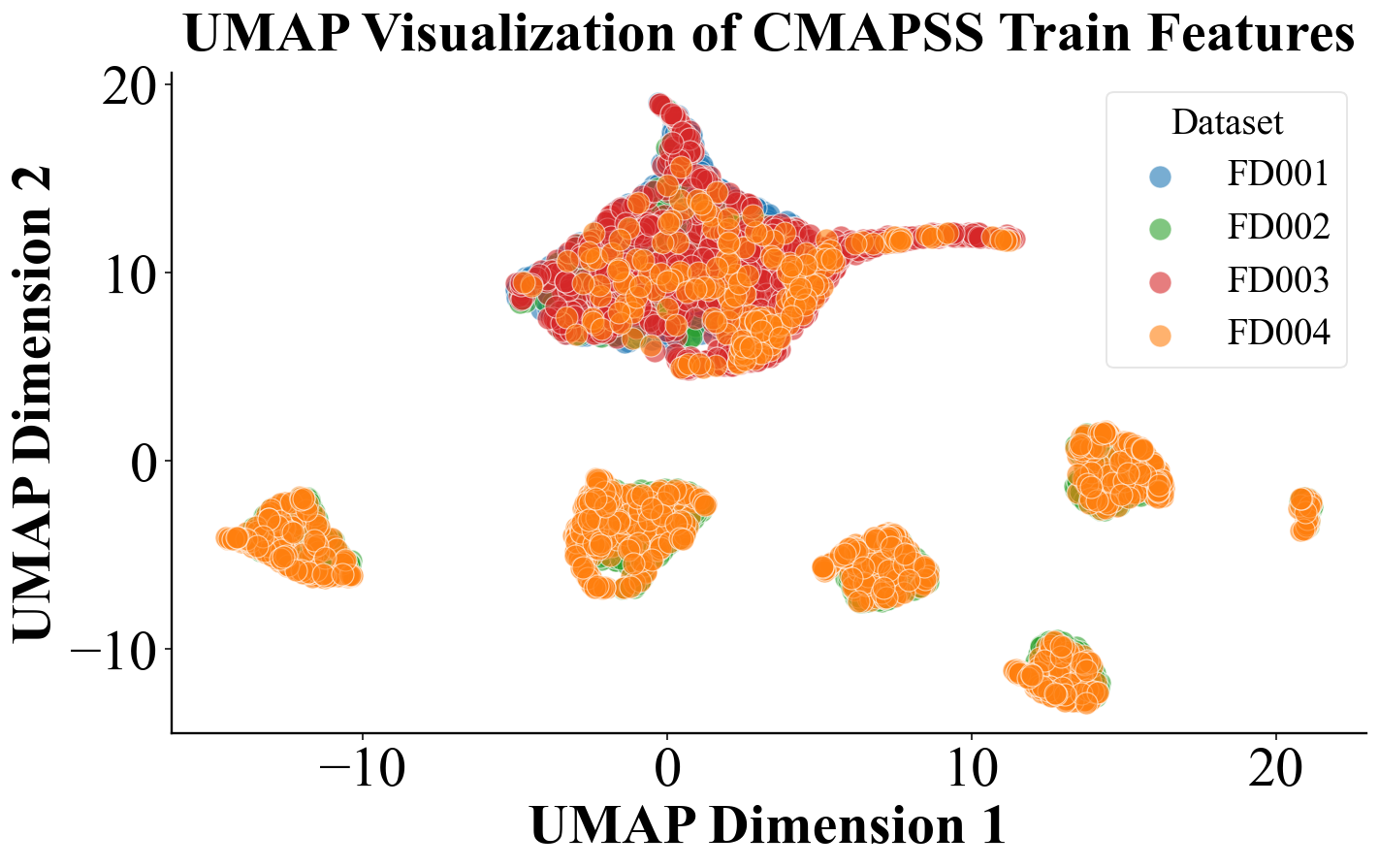}
  \caption{UMAP visualization of C-MAPSS training features across subsets FD001--FD004. Local clustering and separation coexist, reflecting distribution shifts in operating conditions and degradation dynamics. }
  \label{fig:cmapss_umap}
\end{figure}

\begin{figure*}[t]
  \centering
  \includegraphics[width=2\columnwidth]{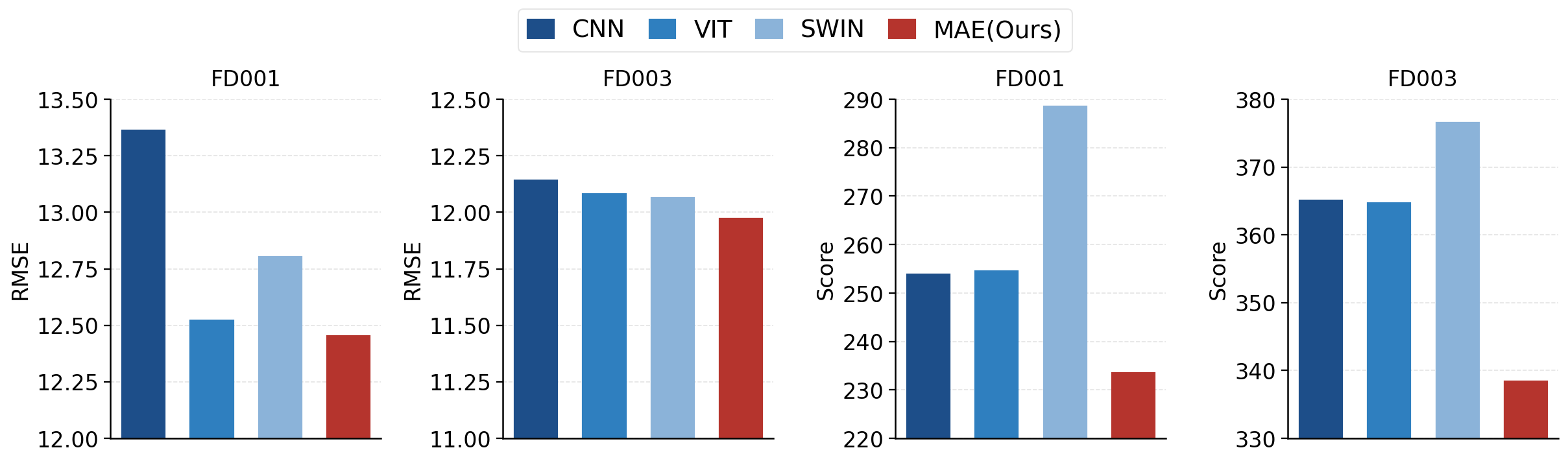}
  \caption{ Comparison of RMSE and Score results on sub-datasets FD001 and FD003 using different visual encoders: a lightweight CNN, ViT, Swin Transformer, and the proposed MAE-based design. }
  \label{fig:rmse_score_fourpanel}
\end{figure*}

\subsection{Model Analysis}

We further conduct an ablation study on the Visual Encoder module. Fig.~\ref{fig:rmse_score_fourpanel} compares different spectrum encoders used to process the converted frequency representations, including a lightweight CNN, a ViT backbone, a Swin Transformer, and our MAE-based design. On both FD001 and FD003, the MAE variant consistently achieves the best results in terms of RMSE and Score. The improvement is particularly evident on Score, where MAE produces markedly lower values than the alternatives, indicating fewer severely penalized early or late predictions.

These results suggest that, given the same frequency-domain input, the choice of Visual Encoder substantially affects the quality of the extracted spectral features. The MAE objective encourages structure-aware representations that better capture degradation-related spectral patterns while being less sensitive to spurious fluctuations, leading to more reliable downstream fusion and regression.

\subsection{Visualization}

\begin{figure*}[t]
	\centering
	\includegraphics[width=2\columnwidth]{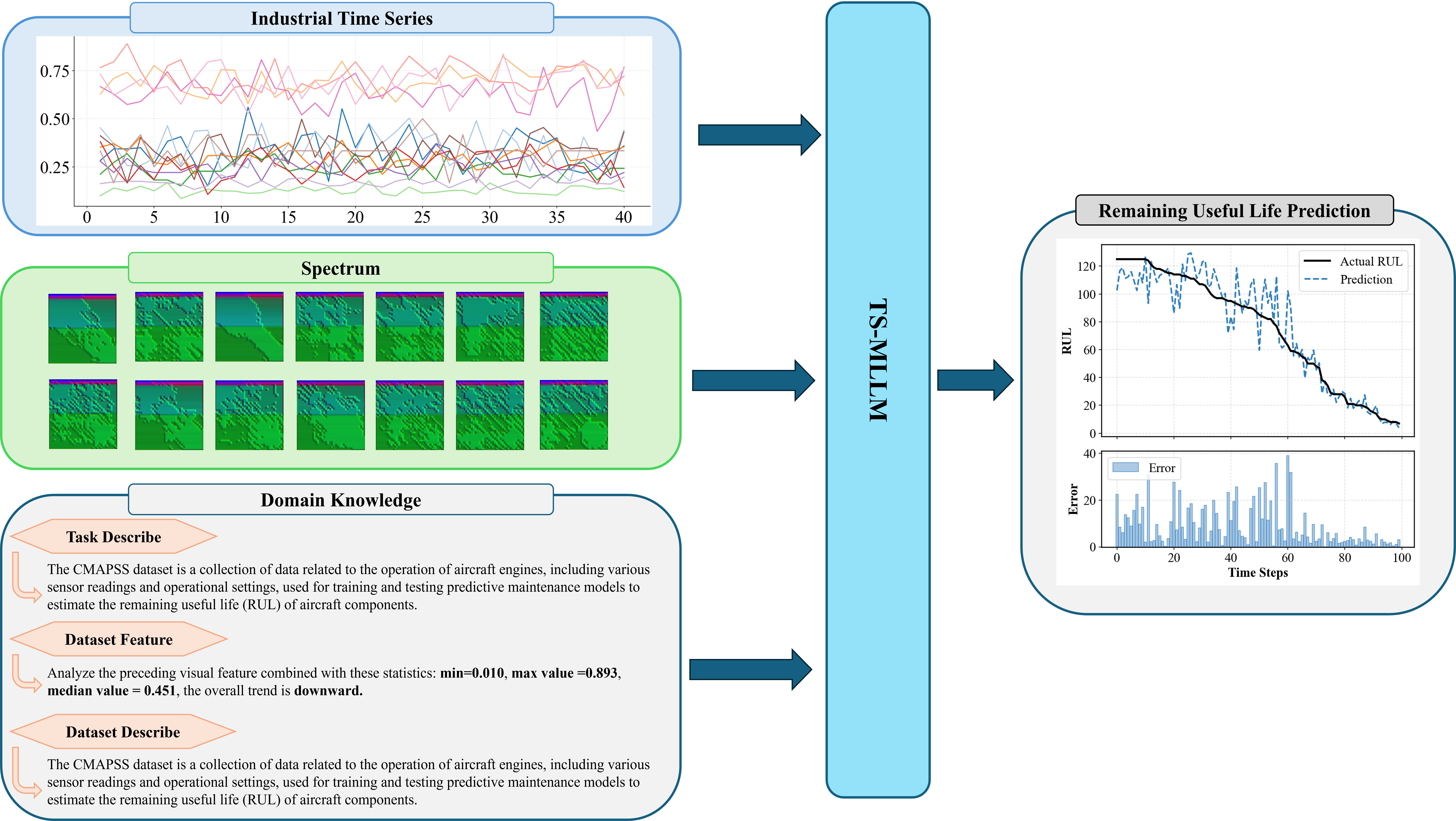}
	\caption{Illustration of TS-MLLM in a representative RUL estimation task. The model integrates industrial time series (sensor readings), spectrum (frequency patterns), and domain knowledge (semantic context) to achieve precise health state assessment. The prediction curve closely tracks the actual degradation process with minimal error throughout the lifecycle.}
	\label{fig:example}
\end{figure*}

To interpret the representation behavior and fusion mechanism of TS-MLLM, we visualize the embedding geometry and the learned fusion weights. Fig.~\ref{fig:fig3_real_umap_grid} shows UMAP projections of the time-series embeddings and the MLLM feature embeddings on FD001 to FD004. Across all subsets, the two types of embeddings form clearly separated clusters with compact intra-cluster structure, indicating that the multi-modal branch encodes information that is not redundant with the temporal branch. The consistent separation across subsets suggests that TS-MLLM preserves modality-specific structure and relies on the fusion module to bridge these complementary cues rather than collapsing them into a single representation.

Fig.~\ref{fig:fig3_real_gate_grid} reports the empirical distributions of gating weights assigned to the time-series features and the MLLM features. The weights span a broad range and vary across subsets, and both branches receive non-trivial weights instead of being dominated by a fixed coefficient. In addition, the presence of both positive and negative values indicates that the gate can either emphasize or suppress a modality depending on the input, supporting the instance-adaptive nature of the temporal-centric fusion.

We further visualize the global distribution of training features on C-MAPSS in Fig.~\ref{fig:cmapss_umap}. The four subsets share a major overlapping region while also exhibiting subset-specific clusters, reflecting distribution shifts induced by different operating conditions and degradation patterns. This observation motivates the need for robust transferable representations, and it is consistent with the design of TS-MLLM, where spectral-aware adaptation provides informative priors and the fusion mechanism selectively integrates multi-modal cues under cross-subset variability.

Finally, Fig.~\ref{fig:example} visualizes a real-case RUL prediction of TS-MLLM using industrial time series, spectrum representations, and domain knowledge inputs. By jointly leveraging temporal degradation dynamics from sensor sequences, frequency-domain signatures captured in the spectrum, and semantic context provided by domain knowledge, TS-MLLM generates the predicted remaining useful life for the target engine.

\section{Conclusion}

This paper proposed TS-MLLM, a unified Multi-modal Large Language Model framework for industrial time-series understanding. By jointly modeling temporal, spectral, and textual modalities, TS-MLLM effectively captures complementary information across domains. The proposed industrial time-series patch modeling, spectrum-aware vision-language model adaptation, and temporal-centric multi-modal fusion mechanisms enable deep temporal representation learning and cross-modal alignment. Extensive experiments on multiple industrial benchmarks confirm the superior robustness and generalization of TS-MLLM, especially in few-shot and complex scenarios. Future work will focus on integrating physics-informed priors and developing foundation models for industrial multi-modal intelligence.

\bibliographystyle{ieeetr}
\bibliography{reference}

@article{zhou2023one,
  title={One fits all: Power general time series analysis by pretrained lm},
  author={Zhou, Tian and Niu, Peisong and Sun, Liang and Jin, Rong and others},
  journal={Advances in neural information processing systems},
  volume={36},
  pages={43322--43355},
  year={2023}
}

@inproceedings{jin2024time,
  title={Time-LLM: Time Series Forecasting by Reprogramming Large Language Models},
  author={Jin, Ming and Wang, Shiyu and Ma, Lintao and Chu, Zhixuan and Zhang, James and Shi, Xiaoming and Chen, Pin-Yu and Liang, Yuxuan and Li, Yuan-fang and Pan, Shirui and others},
  booktitle={International Conference on Learning Representations},
  year={2024}
}

@article{xue2023promptcast,
  title={Promptcast: A new prompt-based learning paradigm for time series forecasting},
  author={Xue, Hao and Salim, Flora D},
  journal={IEEE Transactions on Knowledge and Data Engineering},
  volume={36},
  number={11},
  pages={6851--6864},
  year={2023},
  publisher={IEEE}
}

@article{ansari2024chronos,
  title={Chronos: Learning the Language of Time Series},
  author={Ansari, Abdul Fatir and Stella, Lorenzo and Turkmen, Caner and Zhang, Xiyuan and Mercado, Pedro and Shen, Huibin and Shchur, Oleksandr and Rangapuram, Syama Syndar and Pineda Arango, Sebastian and Kapoor, Shubham and Zschiegner, Jasper and Maddix, Danielle C. and Mahoney, Michael W. and Torkkola, Kari and Gordon Wilson, Andrew and Bohlke-Schneider, Michael and Wang, Yuyang},
  journal={Transactions on Machine Learning Research},
  issn={2835-8856},
  year={2024},
  url={https://openreview.net/forum?id=gerNCVqqtR}
}

@inproceedings{das2024decoder,
  title={A decoder-only foundation model for time-series forecasting},
  author={Das, Abhimanyu and Kong, Weihao and Sen, Rajat and Zhou, Yichen},
  booktitle={Forty-first International Conference on Machine Learning},
year={2024},
}

@article{lan2025gem,
  title={Gem: Empowering mllm for grounded ecg understanding with time series and images},
  author={Lan, Xiang and Wu, Feng and He, Kai and Zhao, Qinghao and Hong, Shenda and Feng, Mengling},
  journal={The annual Neural Information Processing Systems conference (NIPS)},
  year={2025}
}

@article{chen2025domain,
  title={A Domain Knowledge-Guided Industrial Large Model Framework: A Case Study in Battery Health Estimation and Recycling},
  author={Chen, Bingyang and Shao, Haidong and Qin, Yao and Jin, Yang and Hu, Xinming},
  journal={IEEE Transactions on Industrial Informatics},
  year={2025},
  publisher={IEEE}
}

@article{hurst2024gpt,
  title={Gpt-4o system card},
  author={Hurst, Aaron and Lerer, Adam and Goucher, Adam P and Perelman, Adam and Ramesh, Aditya and Clark, Aidan and Ostrow, AJ and Welihinda, Akila and Hayes, Alan and Radford, Alec and others},
  journal={arXiv preprint arXiv:2410.21276},
  year={2024}
}

@article{guo2025deepseek,
  title={Deepseek-r1 incentivizes reasoning in llms through reinforcement learning},
  author={Guo, Daya and Yang, Dejian and Zhang, Haowei and Song, Junxiao and Wang, Peiyi and Zhu, Qihao and Xu, Runxin and Zhang, Ruoyu and Ma, Shirong and Bi, Xiao and others},
  journal={Nature},
  volume={645},
  number={8081},
  pages={633--638},
  year={2025},
  publisher={Nature Publishing Group UK London}
}

@article{zhang2023data,
  title={A data-model interactive remaining useful life prediction approach of lithium-ion batteries based on PF-BiGRU-TSAM},
  author={Zhang, Jiusi and Huang, Congsheng and Chow, Mo-Yuen and Li, Xiang and Tian, Jilun and Luo, Hao and Yin, Shen},
  journal={IEEE Transactions on Industrial Informatics},
  volume={20},
  number={2},
  pages={1144--1154},
  year={2023},
  publisher={IEEE}
}

@article{jin2023adaptive,
  title={An adaptive and dynamical neural network for machine remaining useful life prediction},
  author={Jin, Ruibing and Zhou, Duo and Wu, Min and Li, Xiaoli and Chen, Zhenghua},
  journal={IEEE Transactions on Industrial Informatics},
  volume={20},
  number={2},
  pages={1093--1102},
  year={2023},
  publisher={IEEE}
}

@ARTICLE{ren2024mts,
  author={Ren, Lei and Wang, Haiteng and Laili, Yuanjun},
  journal={IEEE Transactions on Cybernetics}, 
  title={Diff-MTS: Temporal-Augmented Conditional Diffusion-Based AIGC for Industrial Time Series Toward the Large Model Era}, 
  year={2024},
  volume={54},
  number={12},
  pages={7187-7197},
  doi={10.1109/TCYB.2024.3462500}}

@article{zhou2021novel,
  title={A novel soft sensor modeling approach based on difference-LSTM for complex industrial process},
  author={Zhou, Jiayi and Wang, Xiaoli and Yang, Chunhua and Xiong, Wei},
  journal={IEEE Transactions on Industrial Informatics},
  volume={18},
  number={5},
  pages={2955--2964},
  year={2021},
  publisher={IEEE}
}

@article{wang2025meta,
  title={MetaIndux-TS: Frequency-Aware AIGC Foundation Model for Industrial Time Series},
  author={Wang, Haiteng and Ren, Lei and Li, Yikang},
  journal={IEEE Transactions on Neural Networks and Learning Systems},
  pages={1--13},
  year={2021},
  publisher={IEEE}
}

@ARTICLE{ren2023dynamic,
  author={Ren, Lei and Wang, Haiteng and Huang, Gao},
  journal={IEEE Transactions on Neural Networks and Learning Systems}, 
  title={DLformer: A Dynamic Length Transformer-Based Network for Efficient Feature Representation in Remaining Useful Life Prediction}, 
  year={2024},
  volume={35},
  number={5},
  pages={5942-5952},
  doi={10.1109/TNNLS.2023.3257038}}

@inproceedings{yu2023harnessing,
  title={Harnessing LLMs for Temporal Data-A Study on Explainable Financial Time Series Forecasting},
  author={Yu, Xinli and Chen, Zheng and Lu, Yanbin},
  booktitle={Proceedings of the 2023 Conference on Empirical Methods in Natural Language Processing: Industry Track},
  pages={739--753},
  year={2023}
}

@article{wang2025diagllm,
  title={DiagLLM: multimodal reasoning with large language model for explainable bearing fault diagnosis},
  author={Wang, Jie and Li, Tianrui and Yang, Yan and Chen, Shiqian and Zhai, Wanming},
  journal={Science China Information Sciences},
  volume={68},
  number={6},
  pages={160103},
  year={2025},
  publisher={Springer}
}

@article{ouyang2024combined,
  title={Combined meta-learning with CNN-LSTM algorithms for state-of-health estimation of lithium-ion battery},
  author={Ouyang, Tiancheng and Su, Yingying and Wang, Chengchao and Jin, Song},
  journal={IEEE Transactions on Power Electronics},
  volume={39},
  number={8},
  pages={10106--10117},
  year={2024},
  publisher={IEEE}
}

@article{yuan2025lithium,
  title={A lithium-ion battery state of health estimation method utilizing convolutional neural networks and bidirectional long short-term memory with attention mechanisms for collaborative defense against false data injection cyber-attacks},
  author={Yuan, Tianqing and Gao, Feng and Bai, Jing and Sun, Hao},
  journal={Journal of Power Sources},
  volume={631},
  pages={236193},
  year={2025},
  publisher={Elsevier}
}

@article{feng2022spatial,
  title={Spatial-attention and demographic-augmented generative adversarial imputation network for population health data reconstruction},
  author={Feng, Yujie and Wang, Jiangtao and Wang, Yasha and Chu, Xu},
  journal={IEEE Transactions on Big Data},
  volume={9},
  number={4},
  pages={1057--1070},
  year={2022},
  publisher={IEEE}
}

@article{hussain2024big,
  title={Big data analysis for industrial activity recognition using attention-inspired sequential temporal convolution network},
  author={Hussain, Altaf and Hussain, Tanveer and Ullah, Waseem and Khan, Samee Ullah and Kim, Min Je and Muhammad, Khan and Del Ser, Javier and Baik, Sung Wook},
  journal={IEEE Transactions on Big Data},
  year={2024},
  publisher={IEEE}
}

@article{liang2023survey,
  title={A survey on spatio-temporal big data analytics ecosystem: Resource management, processing platform, and applications},
  author={Liang, Huanghuang and Zhang, Zheng and Hu, Chuang and Gong, Yili and Cheng, Dazhao},
  journal={IEEE Transactions on Big Data},
  volume={10},
  number={2},
  pages={174--193},
  year={2023},
  publisher={IEEE}
}

@inproceedings{chen2025visionts,
  title={VisionTS: Visual Masked Autoencoders Are Free-Lunch Zero-Shot Time Series Forecasters},
  author={Chen, Mouxiang and Shen, Lefei and Li, Zhuo and Wang, Xiaoyun Joy and Sun, Jianling and Liu, Chenghao},
  booktitle={Forty-second International Conference on Machine Learning},
year={2025},
}

@article{ding2024alad,
  title={ALAD: A New Unsupervised Time Series Anomaly Detection Paradigm based on Activation Learning},
  author={Ding, Fengqian and Li, Bo and Ben, Xianye and Zhao, Jia and Zhou, Hongchao},
  journal={IEEE Transactions on Big Data},
  year={2024},
  publisher={IEEE}
}

@article{zhong2025patchad,
  title={PatchAD: A lightweight patch-based MLP-mixer for time series anomaly detection},
  author={Zhong, Zhijie and Yu, Zhiwen and Yang, Yiyuan and Wang, Weizheng and Yang, Kaixiang and Chen, CL Philip},
  journal={IEEE Transactions on Big Data},
  year={2025},
  publisher={IEEE}
}

@article{jin2022time,
  title={A time series transformer based method for the rotating machinery fault diagnosis},
  author={Jin, Yuhong and Hou, Lei and Chen, Yushu},
  journal={Neurocomputing},
  volume={494},
  pages={379--395},
  year={2022},
  publisher={Elsevier}
}

@article{lian2024universal,
  title={A universal and efficient multi-modal smart contract vulnerability detection framework for big data},
  author={Lian, Wenjuan and Bao, Zikang and Zhang, Xinze and Jia, Bin and Zhang, Yang},
  journal={IEEE Transactions on Big Data},
  volume={11},
  number={1},
  pages={190--207},
  year={2024},
  publisher={IEEE}
}

@article{shah2024novel,
  title={A novel positional encoded attention-based Long short-term memory network for state of charge estimation of lithium-ion battery},
  author={Shah, Syed Abbas Ali and Niazi, Sajawal Gul and Deng, Shangqi and Azam, Hafiz Muhammad Hamza and Yasir, Khalil Mian Muhammad and Kumar, Jay and Xu, Ziqiang and Wu, Mengqiang},
  journal={Journal of Power Sources},
  volume={590},
  pages={233788},
  year={2024},
  publisher={Elsevier}
}

@article{wang2024timexer,
  title={Timexer: Empowering transformers for time series forecasting with exogenous variables},
  author={Wang, Yuxuan and Wu, Haixu and Dong, Jiaxiang and Qin, Guo and Zhang, Haoran and Liu, Yong and Qiu, Yunzhong and Wang, Jianmin and Long, Mingsheng},
  journal={Advances in Neural Information Processing Systems},
  volume={37},
  pages={469--498},
  year={2024}
}

@article{goswami2024moment,
  title={Moment: A family of open time-series foundation models},
  author={Goswami, Mononito and Szafer, Konrad and Choudhry, Arjun and Cai, Yifu and Li, Shuo and Dubrawski, Artur},
  journal={arXiv preprint arXiv:2402.03885},
  year={2024}
}

@inproceedings{rasul2023lag,
  title={Lag-llama: Towards foundation models for time series forecasting},
  author={Rasul, Kashif and Ashok, Arjun and Williams, Andrew Robert and Khorasani, Arian and Adamopoulos, George and Bhagwatkar, Rishika and Bilo{\v{s}}, Marin and Ghonia, Hena and Hassen, Nadhir and Schneider, Anderson and others},
  booktitle={R0-FoMo: Robustness of Few-shot and Zero-shot Learning in Large Foundation Models},
  year={2023}
}

@inproceedings{yuan2024unist,
  title={Unist: A prompt-empowered universal model for urban spatio-temporal prediction},
  author={Yuan, Yuan and Ding, Jingtao and Feng, Jie and Jin, Depeng and Li, Yong},
  booktitle={Proceedings of the 30th ACM SIGKDD Conference on Knowledge Discovery and Data Mining},
  pages={4095--4106},
  year={2024}
}

@article{ren2024dlformer,
  title={DLformer: A dynamic length transformer-based network for efficient feature representation in remaining useful life prediction},
  author={Ren, Lei and Wang, Haiteng and Huang, Gao},
  journal={IEEE transactions on neural networks and learning systems},
  volume={35},
  number={5},
  pages={5942--5952},
  year={2024},
  publisher={IEEE}
}

@article{xu2023multi,
  author={Xu, Tiantian and Han, Guangjie and Zhu, Hongbo and Taleb, Tarik and Peng, Jinlin},
  journal={IEEE Transactions on Vehicular Technology}, 
  title={Multi-Resolution LSTM-Based Prediction Model for Remaining Useful Life of Aero-Engine}, 
  year={2024},
  volume={73},
  number={2},
  pages={1931-1941},
  doi={10.1109/TVT.2023.3319377}}

@inproceedings{saxena2008damage,
  title={Damage propagation modeling for aircraft engine run-to-failure simulation},
  author={Saxena, Abhinav and Goebel, Kai and Simon, Don and Eklund, Neil},
  booktitle={2008 international conference on prognostics and health management},
  pages={1--9},
  year={2008},
  organization={IEEE}
}

@article{yang2025qwen3,
  title={Qwen3 technical report},
  author={Yang, An and Li, Anfeng and Yang, Baosong and Zhang, Beichen and Hui, Binyuan and Zheng, Bo and Yu, Bowen and Gao, Chang and Huang, Chengen and Lv, Chenxu and others},
  journal={arXiv preprint arXiv:2505.09388},
  year={2025}
}

@ARTICLE{Wang2025amr,
  author={Wang, Haiteng and Ren, Lei and Zhao, Tuo},
  journal={IEEE Transactions on Industrial Informatics}, 
  title={AMR-Net: Adaptive Temporal-Channel Multiresolution Network for Industrial Time-Series Prediction}, 
  year={2025},
  volume={},
  number={},
  pages={1-12},
  doi={10.1109/TII.2025.3639733}}

@article{ZHOU2023109357,
title = {Deep learning-based intelligent multilevel predictive maintenance framework considering comprehensive cost},
journal = {Reliability Engineering \& System Safety},
volume = {237},
pages = {109357},
year = {2023},
issn = {0951-8320},
doi = {https://doi.org/10.1016/j.ress.2023.109357},
url = {https://www.sciencedirect.com/science/article/pii/S0951832023002715},
author = {Kai-Li Zhou and De-Jun Cheng and Han-Bing Zhang and Zhong-tai Hu and Chun-Yan Zhang},
}

@inproceedings{xin2020deebert,
  title={DeeBERT: Dynamic Early Exiting for Accelerating BERT Inference},
  author={Xin, Ji and Tang, Raphael and Lee, Jaejun and Yu, Yaoliang and Lin, Jimmy},
  booktitle={Proceedings of the 58th Annual Meeting of the Association for Computational Linguistics},
  pages={2246--2251},
  year={2020}
}

\begin{IEEEbiography}[{\includegraphics[width=1in,height=2.5in,clip,keepaspectratio]{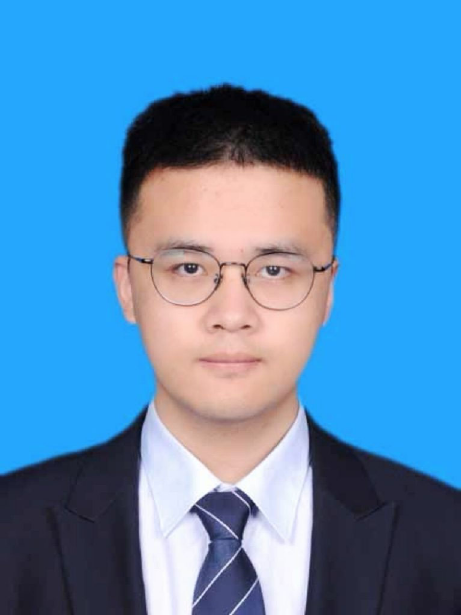}}]{Haiteng Wang} (Graduate Student Member, IEEE) received the B.Eng. Degree in automation engineering from Beihang University, Beijing, China, in 2021, where he is currently working toward the Ph.D. degree in Pattern Recognition and Intelligent Systems with the School of Automation Science and Electrical Engineering. His current research interests include Time Series Prediction, Industrial Foundation Model, Generative Models.
\end{IEEEbiography}

\begin{IEEEbiography}[{\includegraphics[width=1in,height=2.5in,clip,keepaspectratio]{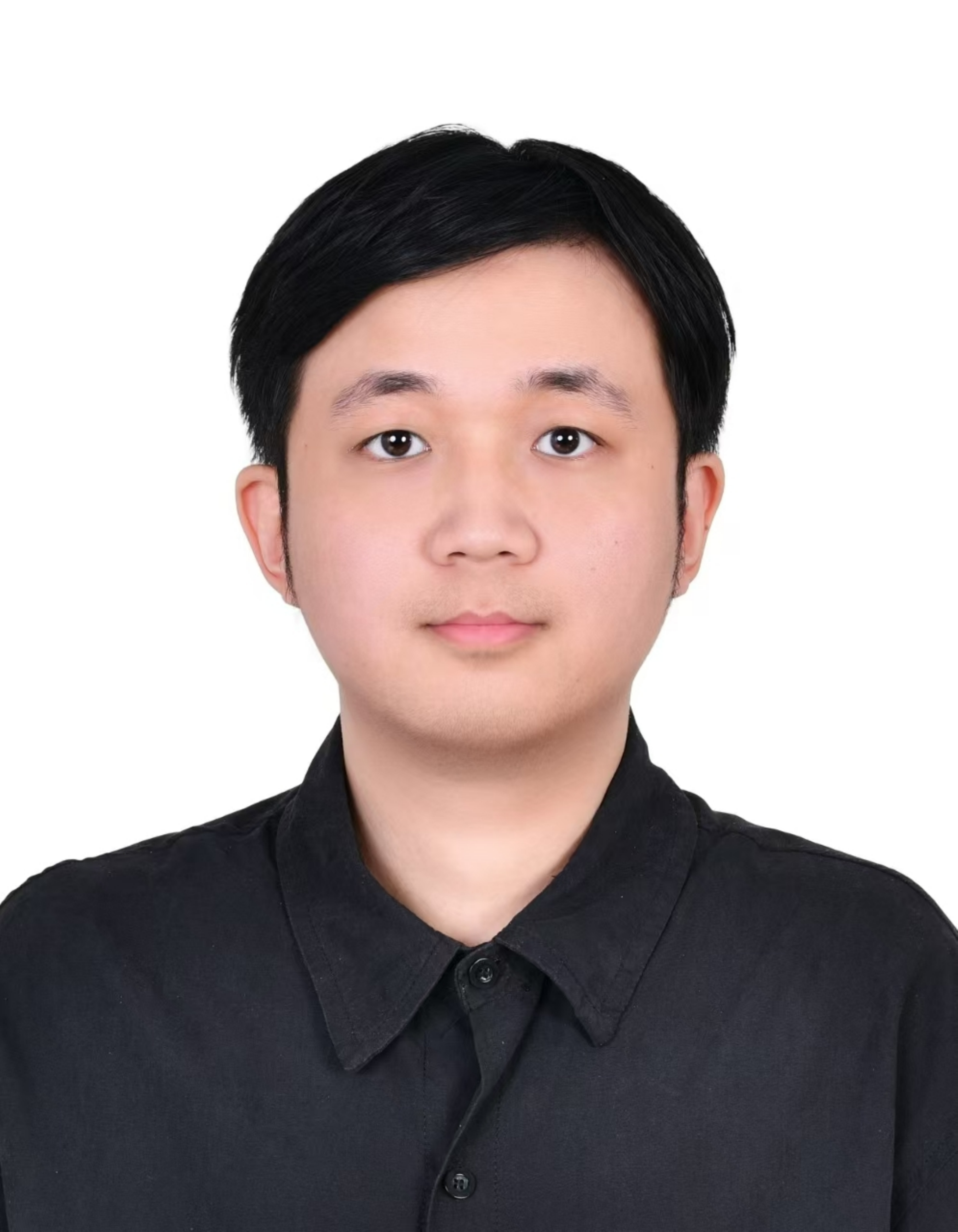}}]{Yikang Li}(Student Member, IEEE) received the B.Eng. degree in automation engineering from Beihang University, Beijing, China, in 2024, where he is currently pursuing the M.S. degree in Control Engineering with the School of Automation Science and Electrical Engineering. His current research interests include Time Series Analysis and Generative Models.
\end{IEEEbiography}

\begin{IEEEbiography}[{\includegraphics[width=1in,height=2.5in,clip,keepaspectratio]{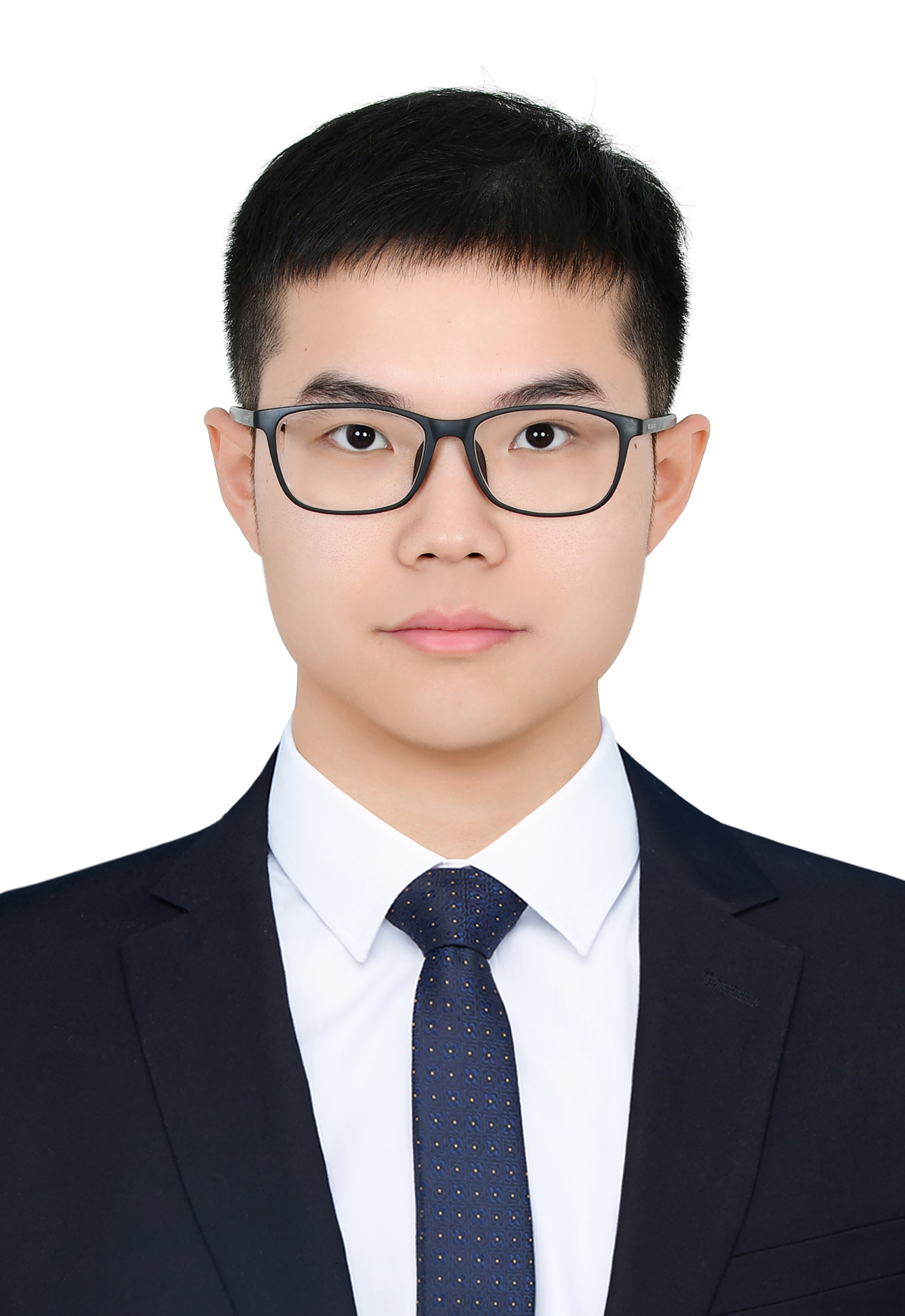}}]{Yunfei Zhu} (Student Member, IEEE) received the B.Eng. degree in Software Engineering from Northeastern University, Shenyang, Liaoning Province, China, in 2025. He is currently pursuing the Ph.D. degree in Software Engineering with the School of Software, Beihang University. His current research interests include Multi-Modal model, Generative Models.
\end{IEEEbiography}

\begin{IEEEbiography}[{\includegraphics[width=1in,height=2.5in,clip,keepaspectratio]{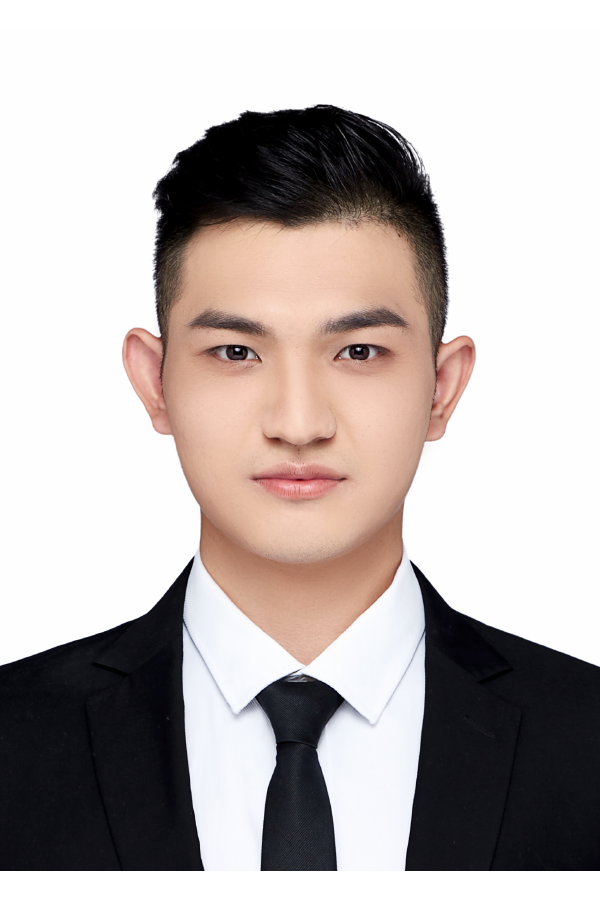}}]{Jingheng Yan} (Graduate Student Member, IEEE) received the B.Eng. Degree in automation from the School of Control Science and Engineering, Shandong University, Jinan, Shandong Province, China, in 2021. He is currently working toward the Ph.D. degree in Control Science and Engineering with the School of Automation Science and Electrical Engineering, Beihang University. His current research interests include Multi-Modal model.
\end{IEEEbiography}

\begin{IEEEbiography}[{\includegraphics[width=1in,height=2.5in,clip,keepaspectratio]{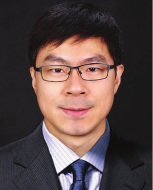}}]{Lei Ren} (Senior Member, IEEE) received the Ph.D. degree in computer science from the Institute of Software, Chinese Academy of Sciences, Beijing, China, in 2009. 
	
He is currently a Professor with the School of Automation Science and Electrical Engineering, Beihang University, Beijing, China, also with the Hangzhou International Innovation Institute, Beihang University, Hangzhou, China, and also with the State Key Laboratory of Intelligent Manufacturing System Technology, Beijing, China. His research interests include neural networks and deep learning, time series analysis, and industrial AI applications. Dr. Ren serves as an Associate Editor for the IEEE Transactions on Neural Networks and Learning Systems, IEEE/ASME Transactions on Mechatronics and other international journals.	
\end{IEEEbiography}

\begin{IEEEbiography}[{\includegraphics[width=1in,height=2.5in,clip,keepaspectratio]{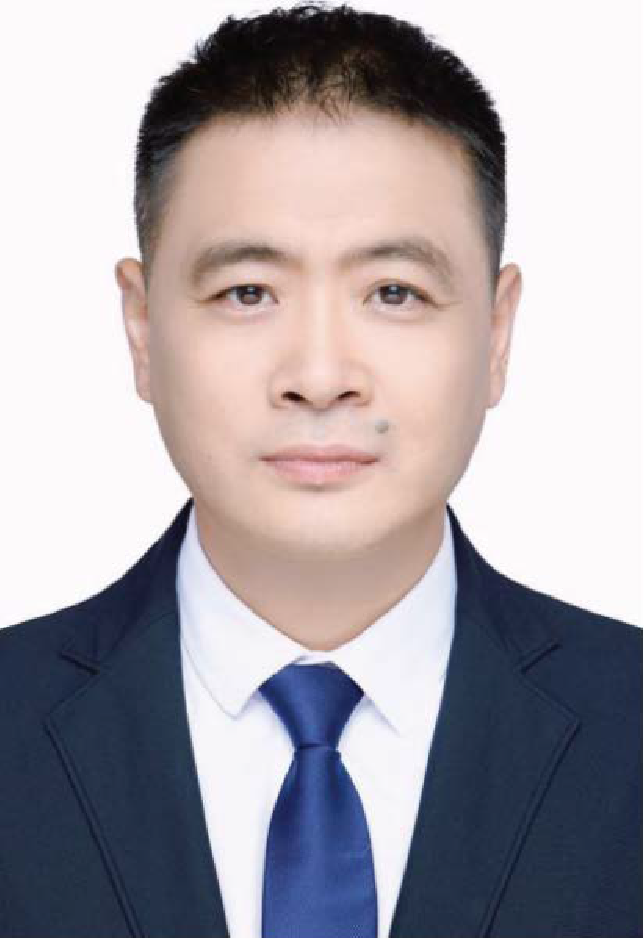}}]{Laurence T. Yang}(Fellow, IEEE) received the B.E. degree in computer science and technology and the B.Sc. degree in applied physics from Tsinghua University, Beijing, China, in 1992, and the Ph.D. degree in computer science from the University of Victoria, Victoria, Canada, in 2006. 
    
He is currently a Professor the School of Computer and Artificial Intelligence, Zhengzhou University, Zhengzhou 450001, China, and also with the Department of Computer Science, St. Francis Xavier University, Antigonish, Canada; and the School of Computer
Science and Technology, Hainan University, Haikou, China. His research interests include parallel and distributed computing, embedded and ubiquitous/pervasive computing, and Cyber-Physical-Social Systems (CPSS).
\end{IEEEbiography}
\end{document}